\documentclass{llncs}
\usepackage{times}
\usepackage{epic}
\usepackage{graphicx}
\usepackage{latexsym}
\usepackage{amsmath}
\usepackage{verbatim}
\usepackage{xspace}

\begin{document}

\newlength{\halftextwidth}
\setlength{\halftextwidth}{0.47\textwidth}
\def\halffigsize{2.2in}
\def\thirdfigsize{1.5in}
\def\negvspace{0in}
\def\posvspace{0em}

\input epsf





\newcommand{\set}{\mathcal}
\newcommand{\myset}[1]{\ensuremath{\mathcal #1}}

\renewcommand{\theenumii}{\alph{enumii}}
\renewcommand{\theenumiii}{\roman{enumiii}}
\newcommand{\figref}[1]{Figure \ref{#1}}
\newcommand{\tref}[1]{Table \ref{#1}}
\renewcommand{\And}{\wedge}
\newcommand{\myldots}{\ldots}

\newtheorem{mydefinition}{Definition}
\newtheorem{mytheorem}{Theorem}
\newtheorem{mylemma}{Lemma}
\spnewtheorem*{myexample}{Running Example}{\bf}{\it}
\spnewtheorem*{myexample2}{Example}{\bf}{\it}
\newtheorem{mytheorem1}{Theorem}
\newcommand{\myproof}{\noindent {\bf Proof:\ \ }}
\newcommand{\myqed}{\mbox{$\Box$}}

\newcommand{\constraint}[1]{\mbox{\sc #1}}
\newcommand{\alldiff}{\constraint{All-Different}\xspace}
\newcommand{\regular}{\constraint{Regular}\xspace}
\newcommand{\grammar}{\constraint{Cfg}\xspace}
\newcommand{\alldiffandsum}{\constraint{All-Different+Sum}\xspace}
\newcommand{\lex}{\constraint{Lex}\xspace}
\newcommand{\among}{\constraint{Among}\xspace}
\newcommand{\gcc}{\constraint{GCC}\xspace}
\newcommand{\egcc}{\constraint{eGCC}\xspace}
\newcommand{\costgcc}{\constraint{Cost-GCC}}
\newcommand{\softgcc}{\constraint{SoftGCC}\xspace}
\newcommand{\minmax}{\constraint{MinMax}}
\newcommand{\dom}{\ensuremath{\mbox{dom}}}
\newcommand{\SwitchC}{\constraint{Switch}}
\newcommand{\MaxSwitch}{\constraint{MaxSwitch}}
\newcommand{\MinSwitch}{\constraint{MinSwitch}}
\newcommand{\sumc}{\constraint{Sum}\xspace}
\newcommand{\interdistance}{\constraint{Inter-Distance}}
\newcommand{\nvalue}{\constraint{NValue}\xspace}
\newcommand{\AtMostNValue}{\constraint{AtMostNValue}\xspace}
\newcommand{\AtLeastNValue}{\constraint{AtLeastNValue}\xspace}
\newcommand{\permutation}{\constraint{Permutation}\xspace}
\newcommand{\mypermutation}{\constraint{Same}\xspace}
\newcommand{\sameset}{\constraint{SameSet}\xspace}
\newcommand{\same}{\constraint{Same}\xspace}
\newcommand{\usedby}{\constraint{Used-By}\xspace}
\newcommand{\uses}{\constraint{Uses}\xspace}
\newcommand{\disjoint}{\constraint{Disjoint}\xspace}
\newcommand{\common}{\constraint{Common}\xspace}
\newcommand{\softalldiff}{\constraint{SoftAllDifferent}\xspace}
\newcommand{\notallequal}{\constraint{Not-All-Equal}\xspace}
\newcommand{\softallequal}{\constraint{Soft-All-Equal}}
\newcommand{\gsc}{\constraint{GSC}}
\newcommand{\sequence}{\constraint{Sequence}\xspace}
\newcommand{\precedence}{\constraint{Precedence}\xspace}
\newcommand{\GCC}{\constraint{GCC}\xspace}
\newcommand{\roots}{\constraint{Roots}\xspace}
\newcommand{\range}{\constraint{Range}\xspace}

\newcommand{\tighter}{\mbox{$\preceq$}}
\newcommand{\stighter}{\mbox{$\prec$}}
\newcommand{\incomparable}{\mbox{$\bowtie$}}
\newcommand{\equivalent}{\mbox{$\equiv$}}

\newcommand{\todo}[1]{{\tt (... #1 ...)}}
\newcommand{\myOmit}[1]{}

\title{Decomposition of the \nvalue constraint}

\author{Christian Bessiere\inst{1}
\and
George Katsirelos\inst{2} \and
Nina Narodytska\inst{3}
\and 
Claude-Guy Quimper\inst{4}
\and
Toby Walsh\inst{3}}
\institute{LIRMM, CNRS, Montpellier, email:
bessiere@lirmm.fr
\and CRIL-CNRS, Lens, email:
gkatsi@gmail.com
\and NICTA and University of NSW,
Sydney, Australia, email: 
\{nina.narodytska,toby.walsh\}@nicta.com.au
\and
Universit\'{e} Laval, email: 
cquimper@gmail.com}

\pdfinfo{
 /Author (Christian Bessiere, George Katsirelos, Nina Narodytska, Claude-Guy Quimper, Toby Walsh)
  /Title (Decomposition of the NValue constraint)
  /Keywords (constraint satisfaction, computational complexity, global constraints)
  /DOCINFO pdfmark}


\maketitle
\begin{abstract}
We study decompositions of the global
\nvalue constraint. Our main contribution
is theoretical: we show that there are 
propagators for global constraints like 
\nvalue  which decomposition 
can simulate with the same time
complexity but with a much greater
space complexity. This suggests that the benefit of a global
propagator may often not be in 
saving time but in saving space. 
Our other theoretical contribution
is to show for the first time
that range consistency can be enforced
on \nvalue with the same worst-case time
complexity as bound consistency. 
Finally, the decompositions we study are
readily encoded as linear inequalities. We
are therefore able to use them in integer
linear programs. 
\end{abstract}

\sloppy
\section{Introduction}

Global constraints are one of the 
distinguishing features of constraint
programming. They capture common 
modelling patterns
and have associated efficient propagators
for pruning the search space. 
For example, \alldiff is one of the best known 
global constraints that has proven
useful in the modelling and solving of
many real world problems.
A number of efficient algorithms have
been proposed to propagate the \alldiff
constraint (e.g. \cite{regin1,Leconte,puget98}).
Whilst there is little debate that
\alldiff is a global
constraint, the formal definition of 
a global constraint is more difficult
to pin down. One property often
associated with global constraints
is that they cannot be decomposed 
into simpler constraints without impacting
either the pruning or the efficiency of propagation
\cite{tobe}.  Recently progress has been
made on the theoretical problem of understanding 
what is and isn't a global constraint. In particular,
whilst a bound consistency propagator
for the \alldiff constraint can be effectively
simulated with a simple decomposition
\cite{bknqwijcai09}, circuit
complexity lower bounds have been used
to prove that a domain
consistency propagator
for \alldiff cannot be polynomially
simulated by a simple decomposition 
\cite{bknwijcai09}.

In this paper, 
we turn to 
a strict generalization of the \alldiff
constraint. \nvalue counts the number
of values used by a set of variables;
the \alldiff constraint ensures that
this count equals the cardinality of the
set. From a theoretical
perspective, the \nvalue constraint
is significantly more difficult to propagate than the \alldiff
constraint since enforcing domain consistency
is known to be NP-hard
\cite{bhhwaaai2004}.
Moreover, as \nvalue is a generalization of \alldiff, 
there exists no polynomial sized decomposition
of \nvalue 
which achieves domain consistency \cite{bknwijcai09}.
Nevertheless, 
we show that decomposition
can simulate the polynomial
time algorithm for enforcing bound consistency on \nvalue
but with a significant space complexity.
We also prove, for the first time,
that range consistency on \nvalue can 
be enforced in the same
worst case time complexity as
bound consistency. 
This contrasts with the \alldiff
constraint where range consistency takes $O(n^2)$ time
\cite{Leconte} but bound consistency
takes just $O(n \log n)$ time \cite{puget98}. 

The main value of these decompositions is theoretical as
their space complexity is
equal to their worst case time complexity. When domains
are large, this space complexity
may be prohibitive. 
In the conclusion, we argue why it appears somewhat inevitable
that the space complexity is equal to the worst case
time complexity. These results suggest
new insight into what is and isn't a global
constraint: a global constraint either provides
more pruning than any polynomial sized decomposition or provides
the same pruning but with lower space complexity.
There are several other theoretical
reasons why the decompositions studied here are interesting. 
First, it is technically interesting that
a complex propagation algorithm like the bound
consistency propagator for \nvalue can be 
simulated by a simple decomposition. 
Second, these decompositions can be readily encoded as 
linear inequalities and used in linear programs. 
In fact, we will report experiments using both constraint and integer
linear programming with these decompositions.
Since global constraints are one of the 
key differentiators between constraint and integer programming,
these decompositions provide us with another
tool to explore the interface between constraint
and integer programming. 
Third, the decompositions give insights 
into how we might add nogood learning
to a \nvalue propagator. 

\section{Background}

A constraint satisfaction problem (CSP) consists of a set of
variables, each with a finite domain of values, and a set of
constraints. 
We use capitals for variables 
and lower case for values. 
We assume values are taken from the set 1 to $d$. 
We write $dom(X_i)$ for the domain of possible values for $X_i$, $min(X_i)$
for the smallest value in $dom(X_i)$,  $max(X_i)$ for the greatest,
and $range(X_i)$ for the interval $[min(X_i),max(X_i)]$.
Constraint solvers typically use
backtracking search to explore the space
of partial assignments. After each assignment,
propagation algorithms prune the search
space by enforcing local consistency properties like 
domain, range  or bound consistency. A constraint is 
\emph{domain consistent} (\emph{DC})
iff when a variable is assigned any of the values in its domain, there
exist compatible values in the domains of all the other variables of
the constraint. Such an assignment is called
a \emph{support}. 
A CSP is domain consistent iff every
constraint is domain consistent.
A constraint is \emph{disentailed} iff 
there is no possible support. 
A propagator which enforces
domain consistency will detect
disentailment, but a propagator
that detects just disentailment will not
enforce domain consistency. 
A constraint is \emph{range consistent} (\emph{RC})
iff, when a variable is
assigned any of the values in its domain, there exist compatible
values between the minimum and maximum domain value
for all the other variables of the constraint.
Such an assignment is called
a \emph{bound support}. 
A constraint is \emph{bound consistent} (\emph{BC})
iff   the minimum and maximum value of every variable of the
constraint belong to a bound support.
A CSP is bound consistent iff every
constraint is bound consistent.
%
We compute the total amortized cost of enforcing a
local consistency down an entire branch of the search tree. 
This captures the incremental cost of propagation.  
Finally, we will assume that 
a propagator is invoked at most once for each domain change and that the
solver uses an optimal propagator to enforce BC on sum
and channeling constraints. Such assumptions hold for modern
solvers like Gecode and Ilog Solver. 
However, we make no assumption about the order of invocation
of the constraints in a decomposition. The upper
bounds we give hold \emph{regardless} of the order in which
constraints are processed.


A \emph{global constraint} is one in which the arity of the
constraint $n$ is a parameter. 
A \emph{decomposition} of a global constraint is a CSP involving 
the $n$ variables of the global constraint (and possibly others),  
involving only  constraints with fixed arity (no global constraint)
or constraints that are themselves decomposable,
such that the size of the CSP is polynomial in the sum of the sizes of
the domains of the $n$ original variables, 
and  such that the
projection of its solutions  on those $n$ variables 
corresponds to the solutions of the global constraint. 
A useful notion is algorithmic
globality \cite{tobe}.
Informally, given a local consistency
property, a global constraint is algorithmically global
if there is no decomposition on which
this local consistency is achieved in the
same time and space complexity.
We suggest here two refinements of
this notion of algorithmic globality.
First, we will separate the space and time
complexity. 
That is, given a local consistency
property, a global constraint is algorithmically global
with respect to time (space) if there is no decomposition on which
this local consistency is achieved in the
same time (space) complexity.
Second, unlike \cite{tobe}, 
we consider decompositions that
may introduce new variables. 
Our results will show that, when we introduce
new variables, \nvalue
is not algorithmically global with respect to time but \emph{is} global with respect to space.

\section{\nvalue constraint}

Pachet and Roy first proposed
the \nvalue constraint 
\cite{pachet1}.
Formally $\nvalue([X_1,\ldots,X_n],N)$ 
ensures that 
$N=|\{X_i \ | \ 1 \leq i \leq n\}|$.
This generalizes several other global 
constraints including \alldiff (which ensures that
the number of values taken by a set of variables equals
the cardinality of the set) and \notallequal 
(which ensures a set of variables take more than one value).
Enforcing domain consistency on the \nvalue
constraint is NP-hard
(Theorem 3 in \cite{bhhwaaai2004}) even when $N$ is
fixed (Theorem 2 in \cite{bhhkwcpaior2005}). 
In fact, just computing the lower bound on $N$ is NP-hard
(Theorem 3 in \cite{bhhwconstraint2007}). 
In addition, enforcing domain consistency on the
\nvalue constraint is not fixed parameter
tractable since it is 
$W$[2]-complete 
\cite{bhhkqwaaai2008}.
However, several polynomial propagation algorithms have
been proposed that achieve bound consistency
and some closely related levels of local
consistency \cite{bcp01,bhhkwcpaior2005,bhhkwconstraint2006}.

\subsection{Simple decomposition}\label{sec:nvalue:simple}

Global constraints can often be decomposed
into simpler, more primitive and small arity constraints.
For example, the \alldiff constraint can
be decomposed into a quadratic number of 
binary inequalities. However, such decomposition
often hinders propagation
and can have a significant 
impact on the solver's ability
to find solutions \cite{swijcai99}. 
We can decompose the \nvalue constraint 
by introducing 0/1 variables
to represent which values are used and posting a sum
constraint on these introduced variables:
\begin{eqnarray}
& X_i=j \rightarrow B_j=1 & \ \ \ \ \forall 1 \leq i \leq n, 1 \leq j \leq d \label{dec1} \\
& B_j=1 \rightarrow \bigvee_{i=1}^n X_i=j  & 
\ \ \ \ \forall 1 \leq j \leq d  \label{dec2}\\
& \sum_{j=1}^d B_j  = N & \label{dec3}
\end{eqnarray}

Note that constraint \ref{dec3} is not a fixed arity constraint, but
can itself be decomposed to ternary sums without hindering
bound propagation.
Unfortunately, this simple decomposition
hinders propagation. 
It can be BC whereas  BC on  the corresponding
\nvalue constraint detects disentailment. 

\begin{mytheorem}
BC on \nvalue is stronger than BC 
on its decomposition into (\ref{dec1}) to (\ref{dec3}).
\end{mytheorem}
\myproof
Clearly BC on \nvalue is at least as strong
as BC on the decomposition. To show strictness,
consider $X_1 \in \{1,2\}$, $X_2 \in \{3,4\}$,
$B_j \in \{0,1\}$ for $1 \leq j \leq 4$, and $N=1$.
Constraints (\ref{dec1}) to (\ref{dec3}) are BC. However, the
corresponding \nvalue constraint has no bound support
and thus enforcing BC on it detects disentailment. 
\myqed

We observe that  enforcing DC instead of BC on constraints
(\ref{dec1}) to (\ref{dec3}) in the example of the proof above  still
does not prune any value.  
To decompose \nvalue 
without hindering propagation, we must look to 
more complex decompositions.

\subsection{Decomposition into \AtMostNValue and \AtLeastNValue}

Our first step in decomposing the \nvalue constraint
is to split it into two
parts: an \AtMostNValue and an \AtLeastNValue constraint.
$\AtLeastNValue([X_1,\ldots,X_n],N)$ holds
iff $N \leq |\{X_i| 1\leq i \leq n\}|$ whilst
$\AtMostNValue([X_1,\ldots,X_n],N)$ holds
iff $|\{X_i| 1\leq i \leq n\}| \leq N$. 

\begin{myexample}
Consider a \nvalue constraint over
the following variables and values:
$$
{\scriptsize
\begin{array}{c|ccccc} 
 & 1 & 2 & 3 & 4 & 5  \\ \hline
X_1 & \ast & \ast & \ast & & \ast  \\ 
X_2 & & \ast & & & \\ 
X_3 & & \ast & \ast & \ast & \\ 
X_4 & & & & \ast & \\
X_5 & & & \ast & \ast & \\
N & \ast & \ast  & & & \ast 
\end{array}
}
$$
Suppose we decompose this into
an \AtMostNValue and an \AtLeastNValue constraint.
Consider the \AtLeastNValue constraint.
The 5 variables can take at most 
4 different values because $X_2,X_3,X_4$, and $X_5$ can only take
values $2, 3$ and $4$. Hence, there is no bound support for
$N=5$. Enforcing BC on the \AtLeastNValue constraint
therefore prunes $N=5$.
Consider now the \AtMostNValue constraint.
Since $X_2$ and $X_4$ guarantee
that we take at least 2 different values,
there is no bound support for $N=1$. 
Hence enforcing BC on an \AtMostNValue constraint
prunes $N=1$.
If $X_1=1$, $3$ or $5$, or $X_5=3$
then any complete assignment uses at least 3 different values. 
Hence there is also no bound support 
for these assignments. Pruning these
values gives bound consistent domains for
the original \nvalue constraint:
$$
{\scriptsize
\begin{array}{c|ccccc} 
 & 1 & 2 & 3 & 4 & 5  \\ \hline
X_1 & & \ast & & &  \\ 
X_2 & & \ast & & & \\ 
X_3 & & \ast & \ast & \ast & \\ 
X_4 & & & & \ast & \\
X_5 & & & & \ast & \\
N & & \ast  & & & 
\end{array}
}
$$
\end{myexample}

To show that decomposing 
the \nvalue constraint into these two 
parts does not hinder propagation in general, 
we will use the following lemma.
Given an assignment $S$ of values, $card(S)$ denotes the number of
distinct values in $S$. Given  a vector of variables  
$X=X_1\ldots X_n$,  
$card_\uparrow(X)=max\{card(S)\mid S\in \Pi_{X_i\in X} range(X_i)\}$ and
$card_\downarrow(X)=min\{card(S)\mid S\in \Pi_{X_i\in X} range(X_i)\}$. 

\begin{mylemma}[adapted from \cite{bhhkwconstraint2006}]
\label{nvalue:prop_1}
Consider $\nvalue([X_1, \ldots, X_n], N)$. If $dom(N) \subseteq [card_\downarrow(X), card_\uparrow(X)]$, 
 then the bounds of $N$ have bound supports
.
\end{mylemma}

\myproof
Let $S_{min}$ be an  assignment of $X$ in $\Pi_{X_i\in X} range(X_i)$ 
with  $card(S_{min}) =
card_\downarrow(X)$  and  $S_{max}$ be an assignment of $X$ in
$\Pi_{X_i\in X} range(X_i)$ 
with  $card(S_{max}) =
card_\uparrow(X)$. Consider the sequence $S_{min}=S_0, S_1,
\ldots,S_n=S_{max}$ where $S_{k+1}$ is the same as $S_k$ except that
$X_{k+1}$ has been assigned its value in $S_{max}$ instead of its
value in $S_{min}$. $|card(S_{k+1})-card(S_k)|\leq 1$
because they only differ on $X_{k+1}$. Hence, 
for any $p\in [card_\downarrow(X),card_\uparrow(X)]$, there
exists $k\in 1..n$ with $card(S_k)=p$.  Thus,  $(S_k,p)$ is a bound
support for $p$ on $\nvalue([X_1, \ldots, X_n], N)$. Therefore, $min(N)$ and
$max(N)$ have a  bound support. 
\myqed

We now prove that
decomposing the \nvalue constraint into
\AtMostNValue and \AtLeastNValue constraints
does not hinder 
pruning when enforcing BC.

\begin{mytheorem}
\label{t:decom_nvalue}
BC on $\nvalue([X_1, \ldots, X_n], N)$ is equivalent to
BC on $\AtMostNValue([X_1, \ldots, X_n], N)$ and  
on $\AtLeastNValue([X_1, \ldots, X_n], N)$. 
\end{mytheorem}
\myproof
  Suppose the \AtMostNValue and \AtLeastNValue constraints
  are BC.  The
  \AtMostNValue constraint guarantees that  $card_\downarrow(X)\leq min(N)$ and the \AtLeastNValue
  constraint guarantees that $card_\uparrow(X) \geq max(N)$.  Therefore, $dom(N) \in
  [card_\downarrow(X),card_\uparrow(X)]$. By Lemma \ref{nvalue:prop_1}, the variable $N$ is bound consistent.

Consider a variable/bound value pair $X_i = b$. Let $(S_{least}^b,p_1)$ be a
bound  support of $X_i = b$ in  the \AtLeastNValue constraint and
$(S_{most}^b,p_2)$  be a bound  support of $X_i = b$ in  the \AtMostNValue
constraint. We  have $card(S_{least}^b)\geq p_1$ and
$card(S_{most}^b)\leq p_2$ by definition of \AtLeastNValue and
\AtMostNValue . 
Consider the sequence $S_{least}^b=S^b_0, S^b_1,
\ldots,S^b_n=S_{most}^b$ where $S^b_{k+1}$ is the same as $S^b_k$ except that
$X_{k+1}$ has been assigned its value in $S_{most}^b$ instead of its
value in $S_{least}^b$. $|card(S^b_{k+1})-card(S^b_k)|\leq 1$
because they only differ on $X_{k+1}$. Hence, there 
exists $k\in 1..n$ with $min(p_1,p_2)\leq card(S^b_k)\leq max(p_1,p_2)$.  
We know that $p_1$ and $p_2$ belong to $range(N)$ because they belong
to bound supports. Thus,  $card(S^b_k)\in range(N)$ and
$(S^b_k,card(S^b_k))$ is a bound support for $X_i=b$ on $\nvalue([X_1,
  \ldots, X_n], N)$.  
%
%
  \myqed

When enforcing domain consistency, Bessiere {\it et al.} \cite{bhhkwconstraint2006} noted that decomposing 
the \nvalue constraint into
\AtMostNValue and \AtLeastNValue constraints
does hinder propagation, but only when $dom(N)$ contains 
just $card_\downarrow(X)$ and 
$card_\uparrow(X)$ and there is a gap in the domain
in-between (see Theorem 1 in \cite{bhhkwconstraint2006}
and the discussion that follows). When enforcing BC, 
any such gap in the domain for $N$ is ignored.

\section{\AtMostNValue constraint}\label{sec:atmost}

We now give a decomposition for the \AtMostNValue 
constraint which does not hinder bound consistency propagation.
To decompose the \AtMostNValue constraint, we 
introduce 0/1 variables, $A_{ilu}$ to represent
whether $X_i$ uses a value in the interval
$[l,u]$, and ``pyramid'' variables, $M_{lu}$
with domains $[0, \min\left(u-l+1, n\right)]$ 
which count the number of values
taken inside the interval $[l,u]$. 
To constrain these introduced variables,
we post the following constraints:
\begin{eqnarray}
&  A_{ilu}  = 1 \iff X_i \in [l, u] & \ \ \ \forall \; 1 \leq i \leq n, 1 \leq l \leq u \leq d \label{eqn::firstAtMostNValue} \\
&  A_{ilu}  \leq M_{lu} & \ \ \ \forall \; 1 \leq i \leq n, 1 \leq l \leq u \leq d \label{eqn::lb_AtMostNValue} \\
  & M_{1u}  = M_{1k} + M_{(k+1)u} & \ \ \ \forall \; 1 \leq k < u \leq d \label{eqn::pyram_AtMostNValue}\\
 & M_{1 d}  \leq N & \label{eqn::lastAtMostNValue}
\end{eqnarray}

\begin{myexample}
Consider the decomposition of
an \AtMostNValue constraint over
the following variables and values:
$$
{\scriptsize
\begin{array}{c|ccccc} 
 & 1 & 2 & 3 & 4 & 5  \\ \hline
X_1 & \ast & \ast & \ast & & \ast  \\ 
X_2 & & \ast & & & \\ 
X_3 & & \ast & \ast & \ast & \\ 
X_4 & & & & \ast & \\
X_5 & & & \ast & \ast & \\
N & \ast & \ast  & & & 
\end{array}
}
$$
Observe that we consider that value  5 for $N$ has already been pruned
by \AtLeastNValue, as will be shown in next sections. 
Bound consistency reasoning on the
decomposition will make the following
inferences. As $X_2=2$, 
from \eqref{eqn::firstAtMostNValue} we get $A_{222}=1$.
Hence by \eqref{eqn::lb_AtMostNValue}, $M_{22}=1$.
Similarly, as $X_4=4$, we get $A_{444}=1$ and $M_{44}=1$. 
Now $N\in \{1,2\}$. By \eqref{eqn::lastAtMostNValue}
and \eqref{eqn::pyram_AtMostNValue},
$M_{15} \leq N$,
$M_{15}=M_{14}+M_{55}$, $M_{14}=M_{13}+M_{44}$,
$M_{13}=M_{12}+M_{33}$, $M_{12}=M_{11}+M_{22}$.
Since $M_{22}=M_{44}=1$, we deduce that $N > 1$
and hence $N=2$. This gives
$M_{11}=M_{33}=M_{55}=0$. 
By \eqref{eqn::lb_AtMostNValue}, 
$A_{111}=A_{133}=A_{155}=A_{533}=0$.
Finally, from \eqref{eqn::firstAtMostNValue},
we get $X_1=2$ and $X_5=3$. This gives us bound consistent
domains for the \AtMostNValue constraint.
\end{myexample}

We now prove that this decomposition 
does not hinder propagation in general. 

\begin{mytheorem}
\label{thm:atmost-bc}
  BC on constraints (\ref{eqn::firstAtMostNValue})
  to (\ref{eqn::lastAtMostNValue}) is equivalent to BC on 
  \AtMostNValue$([X_1,\ldots,X_n], N)$, and takes $O(nd^3)$ time to enforce
down the branch of the search tree. 
\end{mytheorem}
\myproof
\sloppy
First note that changing the domains of the $X$ variables cannot
affect the upper bound of $N$ by the \AtMostNValue constraint and,
conversely, changing the lower bound of $N$ cannot affect the
domains of the $X$ variables. 
  
Let $Y = \{X_{p_1}, \ldots, X_{p_k}\}$ be a maximum cardinality 
subset of variables of $X$ whose ranges are pairwise disjoint (i.e.,
$range(X_{p_i})\cap range(X_{p_j})=\emptyset,\forall i,j\in 1..k,i\neq
j$). Let 
$I_Y = \{[b_i,c_i]\mid b_i = min(X_{p_i}), \ c_i = max(X_{p_i}),
X_{p_i} \in Y\}$ be the corresponding ordered set of disjoint ranges
of the variables in $Y$.  
It has been shown in~\cite{bhhkwcpaior2005} that $|Y|= card_{\downarrow}(X)$. 

Consider the interval $[b_i,c_i] \in I_Y$. Constraints (\ref{eqn::lb_AtMostNValue}) ensure that the variables 
$M_{b_i c_i}$ $i=[1,\ldots,k]$ are greater than or equal to $1$ 
and constraints (\ref{eqn::pyram_AtMostNValue}) ensure that the variable
$M_{1 d}$ is greater than or equal to the sum of lower bounds
of variables $M_{b_i c_i}$, $i=[1,\ldots,k]$, because intervals
$[b_i,c_i]$ are disjoint. 
Therefore, the variable $N$ is greater than or equal to   $card_\downarrow(X)$ and it is bound consistent.

We show that when $N$ is BC and $dom(N)\neq \{card_\downarrow(X)\}$, all
$X$ variables are $BC$. 
Take any assignment $S\in \Pi_{X_i\in X} range(X_i)$ 
such that $card(S)=card_\downarrow(X)$. Let
$S[X_i\gets b]$ be the assignment $S$ where the value of $X_i$ in $S$
has been replaced by $b$, one of the bounds of $X_i$. 
We know that $card(S[X_i\gets b])\in [card(S)-1, card(S)+1]
=[card_\downarrow(X)-1, card_\downarrow(X)+1]$ because
only one variable has been flipped. Hence, any assignment 
$(S,p)$ with $p\geq card_\downarrow(X)+1$ is a bound support. $dom(N)$
necessarily contains such a value $p$ by assumption. 

The only case
when pruning might occur is if the variable $N$ is ground and
$card_\downarrow(X) = N$.  
Constraints (\ref{eqn::pyram_AtMostNValue}) imply that $M_{1d}$ equals the sum of variables
$M_{1,b_1-1} + M_{b_1,c_1} + M_{c_1+1,b_2-1} \ldots + M_{b_N,c_N} + M_{c_N+1,d}$. 
The lower bound of the variable $M_{c_i,b_i}$ is greater than one and
there are $|Y| =card_\downarrow(X)= N$ of these  
intervals. Therefore, by constraint \eqref{eqn::lastAtMostNValue}, the upper bound of variables $M_{c_{i-1}+1,b_i -1}$
that correspond to intervals outside the set $I_Y$ are forced to zero.

%
%
%

  There are $O(nd^2)$ constraints (\ref{eqn::firstAtMostNValue}) and
  constraints (\ref{eqn::lb_AtMostNValue}) that can be
  woken $O(d)$ times down the branch of the search tree. Each requires
    $O(1)$ time for 
  a  total of $O(nd^3)$
  down the branch. 
  There are $O(d^2)$ constraints
  (\ref{eqn::pyram_AtMostNValue}) which can be woken $O(n)$ times  down the
  branch and each invocation takes $O(1)$ time. 
  This gives  a  total of $O(nd^2)$.  
  The final complexity down
  the branch of the search tree is therefore $O(nd^3)$. 
\myqed

The proof of theorem \ref{thm:atmost-bc} also provides the corollary
that enforcing range on consistency on
constraints~\ref{eqn::firstAtMostNValue} enforces range consistency
on \AtMostNValue.
Note that theorem~\ref{thm:atmost-bc} shows that the BC propagator of
\AtMostNValue~\cite{bcp01} is not algorithmically global with respect
to time, as BC can be achieved with a decomposition with comparable
time complexity. On the other hand, the $O(nd^2)$ space complexity of
this decomposition suggests that it is algorithmically global with
respect to space. Of course, we only provide upper bounds here, so it
may be that \AtMostNValue is not algorithmically global with respect
to either time or space.

\section{Faster decompositions}\label{sec:atmost:faster}

We can improve how the solver handles this decomposition
of the \AtMostNValue constraint
by adding implied constraints and by 
implementing specialized propagators.
Our first improvement is to add an implied constraint
and enforce BC on it:
\begin{eqnarray}
M_{1d} & = & \sum_{i=1}^d M_{ii}
\end{eqnarray}
This does not change the asymptotic
complexity of reasoning with the decomposition,
nor does it improve the level 
of propagation achieved.
However, we have found that the fixed point of 
propagation is reached quicker in practice
with such an implied constraint.

Our second improvement decreases the 
asymptotic complexity of enforcing 
BC on the decomposition of Section \ref{sec:atmost}. 
The complexity  is dominated by reasoning with constraints
\eqref{eqn::firstAtMostNValue} which channel
from $X_i$ to $A_{ilu}$ and thence onto $M_{lu}$ (through constraints
\eqref{eqn::lb_AtMostNValue}). 
If  constraints \eqref{eqn::firstAtMostNValue} are not woken
uselessly,  enforcing BC  costs $O(1)$  per constraint down the
branch. Unfortunately, existing
solvers wake up such constraints as soon as a bound is modified, 
thus giving a cost in $O(d)$. 
We therefore implemented
a specialized propagator to channel
between $X_i$ and $M_{lu}$ efficiently. 
To be more precise,
we remove the $O(nd^2)$ variables $A_{ilu}$ and replace them 
with $O(nd)$ Boolean variables $Z_{ij}$.
We then add the following constraints

\begin{align}
  Z_{ij} = 1 \iff &X_i \leq j &
  1 \leq j \leq d
  \label{eq:channel-bounds-Z}\\
  Z_{i(l-1)}=1 \vee Z_{iu} = 0 \; \vee  & \; M_{lu} > 0 &
  1 \leq l \leq u \leq d, 1 \leq i \leq n 
  \label{eq:channel-bounds}
\end{align}

These constraints are enough to channel changes in the bounds of the
$X$ variables to $M_{lu}$. There are $O(nd)$ constraints
\eqref{eq:channel-bounds-Z}, each of which can be propagated
in time $O(d)$ over a branch, for a total of $O(nd^2)$.
There are
$O(nd^2)$ clausal
constraints \eqref{eq:channel-bounds}
and each of them can be made 
BC in time $O(1)$ down a
branch of the search tree, for a total cost of $O(nd^2)$. Since 
channeling
dominates
the asymptotic complexity of the entire decomposition 
of Section \ref{sec:atmost}, this improves
the complexity of this decomposition to $O(nd^2)$.
This is similar to the technique used in \cite{bknqwijcai09} to improve
the asymptotic complexity of the decomposition of the \alldiff constraint.

Our third improvement is to enforce stronger pruning by observing that
when $M_{lu}=0$, we can remove the interval $[l,u]$ from all variables,
regardless of whether this modifies their bounds. This corresponds to  
enforcing RC on constraints \eqref{eqn::firstAtMostNValue}. 
Interestingly, this is sufficient to achieve RC on the \AtMostNValue
constraint. 
Unfortunately,  constraints \eqref{eq:channel-bounds} cannot achieve
this pruning and using constraints \eqref{eqn::firstAtMostNValue} 
increases the complexity of the decomposition back to $O(nd^3)$.
Instead we extend the
decomposition
with $O(d\log d)$ Boolean
variables $B_{il(l+2^k)} \in [0,1], 1 \leq i \leq n, 
1 \leq l \leq d, 0 \leq k \leq \lfloor \log d \rfloor$. 
The 
following constraint ensures that $B_{ijj} = 1 \iff X_i = j$. 

\begin{align}
  {\textrm{\sc DomainBitmap}}(X_i, [B_{i11}, \ldots, B_{idd}])
  \label{eq:DomainBitmap}
\end{align}

Clearly we can enforce RC on this constraint in time $O(d)$ over a
branch, and $O(nd)$ for all variables $X_i$. 
We can then use the following
clausal constraints to channel from variables $M_{lu}$
to these variables and on to the $X$ variables.
These constraints are posted for every $1 \leq i \leq n, 1 \leq l \leq u 
\leq d,
1 \leq j \leq d$ and integers $k$
such that $0 \leq k \leq \lfloor \log d \rfloor$:  
\begin{align}
  B_{ij(j+2^{k+1}-1)} = 1 &\vee B_{ij(j+2^{k}-1)}=0 \label{eq:channel-B-first}\\
  B_{ij(j+2^{k+1}-1)} = 1 & \vee B_{i(j+2^k)(j+2^{k+1}-1)}=0 
  \label{eq:channel-B-last} \\
  M_{lu} \neq 0 &\vee B_{il(l+2^{k}-1)} = 0 
  & 2^k \leq u-l+1 < 2^{k+1}
  \label{eq:channel-range-first}\\
  M_{lu} \neq 0 &\vee B_{i(u-2^k+1)u} = 0
  & 2^k \leq u-l+1 < 2^{k+1}
  \label{eq:channel-range-last}
\end{align}

The variable $B_{il(l+2^k-1)}$, similarly to the variables $A_{lu}$, is
true when  $X_i \in [l, l+2^k-1]$, but instead of having one
such variable for every interval, we only have them for 
intervals whose length is a power of two.
When $M_{lu} = 0$, with $2^k \leq u-l+1 < 2^{k+1}$, 
the constraints
(\ref{eq:channel-range-first})--(\ref{eq:channel-range-last})
set to 0 the 
$B$ variables that correspond to the two intervals of length $2^k$
that start at $l$ and finish at $u$, respectively. In turn,
the constraints
(\ref{eq:channel-B-first})--(\ref{eq:channel-B-last})
set to 0 the $B$ variables that correspond to intervals of 
length $2^{k-1}$, all the way down to intervals of size 1. These trigger
the constraints \eqref{eq:DomainBitmap}, 
so all values in the interval $[l,u]$ are
removed from the domains of all variables.

\begin{myexample2}
  Suppose $X_1 \in [5, 9]$. Then, by \eqref{eq:channel-bounds-Z}, 
  $Z_{14} = 0$, $Z_{19} = 1$ and by \eqref{eq:channel-bounds}, $M_{59}>0$. 
  Conversely, suppose $M_{59} = 0$ and $X_1 \in [1, 10]$. Then, by 
  (\ref{eq:channel-range-first})--(\ref{eq:channel-range-last}),
  we get $B_{158} = 0$ and $B_{169} = 0$. 
  From $B_{158} = 0$ and
  (\ref{eq:channel-B-first})--(\ref{eq:channel-B-last})
  we get $B_{156} = 0$, $B_{178} = 0$, $B_{155} = B_{166} = 
  B_{177} = B_{188} = 0$, and by \eqref{eq:DomainBitmap}, the interval
  $[5,8]$ is pruned from $X_1$. Similarly, $B_{169}=0$ causes
  the interval $[6,9]$ to be removed from $X_1$, so
  $X_1 \in [1,4] \cup \{ 10 \}$.
\end{myexample2}

Note that RC can be enforced on each of these constraints in 
constant time over a branch. There exist $O(nd\log d)$ of the constraints
\eqref{eq:channel-B-first}--\eqref{eq:channel-B-last} 
and $O(nd^2)$ of the
constraints~\eqref{eq:channel-range-first}--\eqref{eq:channel-range-last}, 
so the total time to propagate them all
down a branch is $O(nd^2)$.

\section{\AtLeastNValue constraint}

There is 
a similar decomposition for the
\AtLeastNValue constraint. 
We introduce 0/1 variables, $A_{ilu}$ to represent
whether $X_i$ uses a value in the interval
$[l,u]$, and 
integer variables, $E_{lu}$ with
domains $[0,n]$ to count 
the number of times values in $[l,u]$ are \emph{re}-used, that is, how
much the number of variables taking values in $[l,u]$ exceeds the
number $u-l+1$ of values in $[l,u]$. 
To constrain
these introduced variables, we post the following constraints:
\begin{eqnarray}
  & A_{ilu} = 1 \iff X_i \in [l, u] & \ \ \ \forall \; 1 \leq i \leq n, 1 \leq l \leq u \leq d \label{eqn::atleastnvalue-1} \\
 & E_{lu}  \geq \sum_{i=1}^n A_{ilu} - (u-l+1) & \ \ \ \forall \; 1\leq l \leq u \leq d 
 \label{eqn::atleastnvalue-e}\\
&  E_{1u} = E_{1k} + E_{(k+1)u} & \ \ \ \forall \; 1 \leq k < u \leq d \label{eqn::pyram_AtLeastNValue} \\
&  N\leq n-E_{1 d} \label{eqn::atleastnvalue-2}
\end{eqnarray}
\begin{myexample}
Consider the decomposition of an \AtLeastNValue constraint over
the following variables and values:
$$
{\scriptsize
\begin{array}{c|ccccc} 
 & 1 & 2 & 3 & 4 & 5  \\ \hline
X_1 & \ast & \ast & \ast & & \ast  \\ 
X_2 & & \ast & & & \\ 
X_3 & & \ast & \ast & \ast & \\ 
X_4 & & & & \ast & \\
X_5 & & & \ast & \ast & \\
N & \ast & \ast  & & & \ast 
\end{array}
}
$$
Bound consistency reasoning on the
decomposition will make the following
inferences. As $dom(X_i)\subseteq [2,4]$ for $i\in 2..5$, 
from \eqref{eqn::atleastnvalue-1} we get $A_{i24}=1$ for $i\in 2..5$.
Hence, by \eqref{eqn::atleastnvalue-e}, $E_{24}\geq 1$. 
By \eqref{eqn::pyram_AtLeastNValue}, 
$E_{15}=E_{14}+E_{55}$, $E_{14}=E_{11}+E_{24}$. Since $E_{24}\geq 1$
we deduce that  $E_{15}\geq 1$. 
Finally, from \eqref{eqn::atleastnvalue-2} and the fact that  $n=5$,
we  get $N\leq 4$.  This gives us bound consistent
domains for the \AtLeastNValue constraint.
\end{myexample}

We now prove that this decomposition
does not hinder propagation in general. 

\begin{mytheorem}
  \label{thm:atleastnvalue-bc}
  BC on the constraints
  \eqref{eqn::atleastnvalue-1} 
  to \eqref{eqn::atleastnvalue-2} is equivalent to BC
  on \AtLeastNValue$([X_1,\ldots,X_n], N)$, and takes $O(nd^3)$ time to enforce
down the branch of the search tree.
\end{mytheorem}

\myproof
  First note that changing the domains of the $X$ variables cannot
  affect the lower bound of $N$ by the \AtLeastNValue constraint and,
  conversely, changing the upper bound of $N$ cannot affect the
  domains of the $X$ variables. 
  
  It is known \cite{bcp01} that $card_{\uparrow}(X)$ is equal to the
  size of a maximum matching $M$ in the value graph of the
  constraint. Since $N\leq n-E_{1d}$, we show that the lower bound of
  $E_{1d}$ is equal to $n-|M|$.\footnote{We assume that $E_{1d}$
    is not pruned by other constraints.}
  We first show that we can construct a matching $M(E)$ of size
  $n-min(E_{1d})$, then show that it is a maximum matching. The proof uses 
  a partition of the interval $[1,d]$ into a set of maximal
  saturated intervals $I = \{[b_j,c_j]\}$, $j=1,\ldots,k$ such that $min(E_{b_j, c_j}) = \sum_{i=1}^n min(A_{ib_jc_j}) -
  (c_j-b_j+1)$ and a set of unsaturated intervals $\{[b_j,c_j]\}$such that 
  $min(E_{b_j, c_j}) = 0$.

  Let $I=\{ [b_j, c_j] \mid j \in [1\ldots k] \}$ be the ordered set of maximal
  intervals such that $min(E_{b_j, c_j}) = \sum_{i=1}^n min(A_{ib_jc_j}) -
  (c_j-b_j+1)$. Note that the intervals in $I$ are disjoint otherwise 
  intervals are not maximal. An interval $[b_i,c_i]$ is smaller than
  $[b_j,c_j]$ iff $c_i < b_j$. We denote the union of the first $j$ intervals  $D_I^j = \bigcup_{i=1}^j [b_i,c_i]$,
  $j =[1,\ldots,k]$, $p = |D_I^k|$ and
  the variables whose domain is inside one of intervals $I$ 
  $X_I=\{X_{p_i}| dom(X_{p_i}) \subseteq D_I^k\}$. 
  
  Our construction of a matching uses two sets of
  variables, $X_I$ and $X \setminus X_I$. First, we identify the cardinality of these
  two sets. Namely, we show that the size of the set $X_I$ is  $p + min(E_{1,d})$
  and the size of the set $X\setminus X_I$ is $n - (p + min(E_{1,d}))$.
  
  Intervals $I$ are saturated therefore each value from these intervals are
  taken by a variable in $X_I$. Therefore, $X_I$ has size at least $p$.
  Moreover, there exist $min(E_{1d})$ additional variables that take values from $D_I^k$,
  because values from intervals between two consecutive intervals in $I$ 
  do not contribute to the lower bound of the variable $E$ by construction of $I$.
  Therefore, the number of variables in $D_I^k$ is at least $p + min(E_{1,d})$.
  Note 
  that constraints \eqref{eqn::pyram_AtLeastNValue} imply that $E_{1d}$ equals 
  the sum of variables $E_{1,b_1-1} + E_{b_1,c_1} + E_{c_1+1,b_2-1} \ldots + E_{b_k,c_k} + E_{c_k+1,d}$.
  As intervals in $I$ are disjoint then $\sum_{i=1}^k min(E_{b_i,c_i}) = |X_I| - p$. 
  If $|X_I| > p + min(E_{1,d})$ then  $\sum_{i=1}^k min(E_{b_i,c_i}) >  min(E_{1,d})$ and the lower
  bound of the variable $E_{1d}$ will be increased. Hence, $|X_I| = p + min(E_{1,d})$.
  
  Since all these intervals are saturated, we can
  construct a matching $M_I$ of size $p$ using the variables in
  $X_I$. The size of $X \setminus X_I$ is $n-p-min(E_{1d})$. We show by
  contradiction that we can construct a matching $M_{D-D^k_I}$ of size
  $n-p-min(E_{1d})$ using the variables in $X \setminus X_I$ and the
  values $D-D_I^k$.

  Suppose such a matching does not exist. Then, there exists an
  interval $[b, c]$ such that $|(D \setminus D_I^k) \cap [b,c]| < \sum_{i
    \in X \setminus X_I} min(A_{ibc})$, i.e., after consuming the values in
  $I$ with variables in $X_I$, we are left with fewer values in
  $[b,c]$ than variables whose domain is contained in $[b,c]$.  We
  denote $p' = |[b,c] \cap D^k_I|$,
  so that $p'$ is the number of values inside the interval $[b,c]$
  that are taken by variables in $X_I$. The total number of variables
  inside the interval $[b,c]$ is greater than or equal to
  $\sum_{i=1}^n min(A_{ibc})$. The total number of variables $X_I$ inside
  the interval $[b,c]$ equals to $p' + min(E_{b,c})$. 
  Therefore, $\sum_{i \in  X \setminus X_I} min(A_{ibc}) \leq \sum_{i=1}^n min(A_{ibc}) - p' -
  min(E_{b,c})$. On the other hand, the number of values that are not
  taken by the variables $X_I$ in the interval $[b,c]$ is $c-b+1 -
  p'$. Therefore, we obtain the inequality $c-b+1 -p' < \sum_{i=1}^n
  min(A_{ibc}) - p' - min(E_{b,c})$ or $ min(E_{bc}) < \sum_{i=1}^n min(A_{ibc}) -
  (c-b+1) $.  By construction of $I$, $\sum_{i=1}^n min(A_{ibc}) - (c-b+1)
  < min(E_{bc})$, otherwise the intervals in $I$ that are subsets of
  $[b,c]$ are not maximal. This leads to a contradiction, so we can
  construct a matching $M(E)$ of size $n-min(E_{1d})$.

  Now suppose that $M(E)$ is not a maximum matching.  This means
  that $min(E_{1d})$ is overestimated by propagation on
\eqref{eqn::atleastnvalue-1} 
  and \eqref{eqn::atleastnvalue-2}.  Since
  $M(E)$ is not a maximum matching, there exists an augmenting
  path of $M(E)$, that produces $M'$, such that $|M'| =
  |M(E)|+1$. This new matching covers all the values that
  $M(E)$ covers and one additional value $q$. We show that $q$
  cannot belong to the interval $[1,d]$. 

	The value $q$ cannot be in any   interval in $I$, 
	because all values in $[b_i,c_i] \in I$ are used by variables whose domain is contained in $[b_i,c_i]$. 
	In addition, $q$ cannot be in an interval $[b,c]$ between two consecutive 
	intervals in $I$, because those intervals do
  not contribute to the lower bound of $E_{1d}$. Thus, $M'$ cannot
  cover more values than $M(E)$ and they must have the same size,
  a contradiction.


We show that when $N$ is BC and $dom(N)\neq \{card_\uparrow(X)\}$, all
$X$ variables are $BC$. 
Take any assignment $S\in \Pi_{X_i\in X} range(X_i)$ 
such that $card(S)=card_\uparrow(X)$. Let
$S[X_i\gets b]$ be the assignment $S$ where the value of $X_i$ in $S$
has been replaced by $b$, one of the bounds of $X_i$. 
We know that $card(S[X_i\gets b])\in [card(S)-1, card(S)+1]
=[card_\uparrow(X)-1, card_\uparrow(X)+1]$ because
only one variable has been flipped. Hence, any assignment 
$(S,p)$ with $p\leq card_\uparrow(X)-1$ is a bound support. $dom(N)$
necessarily contains such a value $p$ by assumption. 

  We now show that if $N=card_\uparrow(X)$, enforcing BC on the constraints
 \eqref{eqn::atleastnvalue-1}--\eqref{eqn::atleastnvalue-2} makes the
  variables $X$ BC with respect to the \AtLeastNValue constraint.
We first observe  that in a  bound support, variables  $X$ must take
the maximum number of different values because $N=card_\uparrow(X)$. 
Hence, in a bound support, variables $X$ 
that 
are not included in a saturated interval
 will take values outside any saturated interval they overlap and they
 all take different values. 
We recall that $min(E_{1d})=n-|M|=n-card_\uparrow(X)$. Hence, by
constraint \eqref{eqn::atleastnvalue-2},  $E_{1d}=n-N$. 
  We recall the the size of set $X_I$ equals $ p + E_{1d}$.  
  Constraints \eqref{eqn::pyram_AtLeastNValue} imply that $E_{1d}$ equals 
  the sum of variables $E_{1,b_1-1} + E_{b_1,c_1} + E_{c_1+1,b_2-1} \ldots + E_{b_k,c_k} + E_{c_k+1,d}$
  and $\sum_{i=1}^k min(E_{b_i,c_i}) =|X_I| - p=
  min(E_{1d})=max(E_{1d}) $. Hence, by
  constraints~\eqref{eqn::pyram_AtLeastNValue}, the upper bounds of
  all variables $E_{b_i,c_i}$ that correspond to the saturated intervals
  are forced to  $min(E_{b_i,c_i})$. 
Thus, by constraints \eqref{eqn::atleastnvalue-1} and
\eqref{eqn::atleastnvalue-e}, all variables in $X\setminus X_I$ have
their bounds pruned if they belong to  $D^k_I$. 
By constraints~\eqref{eqn::pyram_AtLeastNValue} again, 
  the upper bounds of all
  variables $E_{lu}$
  that correspond to the  unsaturated intervals are forced to take
  value 0, and all variables $E_{l'u'}$ with $[l',u']\subseteq[l,u]$
  are forced to 0
  as well. 
Thus, by constraints 
\eqref{eqn::atleastnvalue-1} and 
\eqref{eqn::atleastnvalue-e}, all variables in $X\setminus X_I$ have
their bounds pruned  if they belong to a Hall interval of other
variables in $X\setminus X_I$. This is  what BC on the \alldiff
constraint does \cite {bknqwijcai09}. 

  There are $O(nd^2)$ constraints \eqref{eqn::atleastnvalue-1} that can be
  woken $O(d)$ times down the branch of the search tree in  $O(1)$, so
  a  total of $O(nd^3)$
  down the branch. 
  There are $O(d^2)$ constraints
  \eqref{eqn::atleastnvalue-e} which can be 
  propagated in time $O(n)$
  down the
  branch 
  for a $O(nd^2)$.  
  There are $O(d^2)$ constraints
  \eqref{eqn::pyram_AtLeastNValue} which can be woken $O(n)$ times each down the
  branch 
for a total cost in  $O(n)$ time down the
  branch. Thus  a  total of $O(nd^2)$. 
  The final complexity down
  the branch of the search tree is therefore $O(nd^3)$. 
  \myqed

  The complexity of enforcing BC on \AtLeastNValue can
  be improved to $O(nd^2)$ in a way similar to that described in Section
  \ref{sec:atmost:faster} and in \cite{bknqwijcai09}. 
  As with \AtMostNValue, enforcing RC on constraints 
  (\ref{eqn::atleastnvalue-1}) enforces RC on \AtLeastNValue, but
  in this case we cannot reduce the complexity below $O(nd^3)$.
  Similarly to \AtMostNValue, theorem~\ref{thm:atleastnvalue-bc} shows
  that the bound consistency propagator of \AtLeastNValue is not
  algorithmically global with respect to time and provides evidence
  that it is algorithmically global with respect to space.

\section{Experimental results}

As noted before, the main value of these
decompositions is theoretical: 
demonstrating that the bound consistency propagator
of~\cite{bcp01}
for the \nvalue constraint can be simulated
using a simple decomposition with comparable time complexity
over a branch of the search tree
but greater space complexity. 
To see when this space complexity hits, 
we performed some experiments. We used
a benchmark problem, the dominating set of the Queen's graph
used in previous studies of \nvalue\cite{bhhkwconstraint2006}
and ran experiments with
Ilog Solver 6.2 and Ilog CPLEX 9.1 on an Intel Xeon 4 CPU, 2.0 Ghz, 4Gb RAM.  
%
%
The dominating set of the Queen's graph
problem is to put the minimum number of queens on a $n \times n$ chessboard,
so that each square either contains a
queen or  is attacked by one. 
This is equivalent to the dominating set problem of the
Queen's graph. 
Each vertex in the Queen's graph corresponds to a square of the chessboard and there exists
an edge between two vertices iff
a queen from one square can attack a queen from the other square. 
To model the problem, we use a variable $X_i$ for each
square, and values from $1$ to $n^2$ and
post a single $\AtMostNValue([X_1,\ldots, X_{n^2}],N)$ constraint. 
The value $j$ belongs to $dom(X_i)$ iff
there exists an edge  $(i, j)$ in the Queen's graph or $j=i$. 
For $n\leq 120$,
all minimum dominating sets for the Queen's problem 
are either of size $\left\lceil n/2\right\rceil$ or $\left\lceil n/2 + 1\right\rceil$
\cite{ostergard}. We therefore only solved instances for
these two values of $N$. 

We compare our decomposition with the simple decomposition of the $\AtMostNValue$ constraint
in  Ilog Solver and Ilog CPLEX solvers.
The simple decomposition 
is the one described in Section \ref{sec:nvalue:simple} except that in
constraint \eqref{dec3}, we replace ``$=$'' by ``$\leq$''. 
We denote this decomposition $Occs$ and $Occs^{CPLEX}$ in Ilog Solver and CPLEX, respectively.
To encode this decomposition into an integer linear program,
we introduce literals $b_{ij}$, $i,j \in [1,n^2]$ and 
use a direct encoding with $b_{ij}$ for the truth of $X_i = j$ and
channeling inequalities  $1 - b_{ij} + B_{j} \geq 1 $, $i,j \in [1,n^2]$. We use the direct
encoding of variables domains to avoid using logic constraints, like disjunction
and implication constraints in CPLEX. The default transformation
of logic constraints in CPLEX appears to generate
large ILP models and this slows down the search.

%
The BC decomposition is described
in Section \ref{sec:atmost}, which we call
$Pyramid_{BC}$ and $Pyramid_{BC}^{CPLEX}$ in Ilog Solver and CPLEX, respectively. 
%
In Ilog Solver, as explained in Section \ref{sec:atmost:faster},
we channel the variables $X_i$ directly to the pyramid variables $M_{lu}$  to avoid introducing many
auxiliary variables $A_{ilu}$ and we  add the redundant constraint $\sum_{i=1}^{n^2}M_{ii} = M_{1,n^2}$ to the decomposition 
to speed up the propagation across the pyramid. We re-implemented the ternary sum constraint in Ilog
for a 30\% speedup.

To encode the BC decomposition into an integer linear program, we use the linear encoding of variables domains~\cite{osccp07}.
We introduce literals $c_{ij}$ for the truth of $X_i \leq j$,
and the channeling inequalities of the form
$ c_{i(l-1)} + 1 -  c_{iu} + M_{lu} \geq 1 $. 
We again add the redundant constraint $\sum_{i=1}^{n^2}M_{ii} = M_{1,n^2}$. 
Finally, we  post constraints~\eqref{eqn::pyram_AtMostNValue}  as lazy constraints in CLPEX.
Lazy constraints are constraints that are not expected to be violated 
when they are omitted. These constraints are not taken into account 
in the relaxation of the problem and are only included 
when they 
violate an integral solution. 
%

%

\begin{table}[htb]
\begin{center}
{\small
\caption{\label{t:t1} Backtracks and rumtime (in seconds) to
solve the dominating set problem for the Queen's graph. 
}
\begin{tabular}{|  c|c ||rr|rr|rr|rr|}
\hline
$n$ & $N$ 
&\multicolumn {2}{|c|}{$Occs$}
&\multicolumn {2}{|c|}{$Pyramid_{BC}$}  
&\multicolumn {2}{|c|}{$Occs^{CPLEX}$} 
&\multicolumn {2}{|c|}{$Pyramid_{BC}^{CPLEX}$}  \\
\hline 
\multicolumn {2}{|c||}{}
&backtracks & time &backtracks & time &backtracks & time &backtracks & time\\
\hline   

 5 & 3 &       34 &  0.01  
&         7 &  \textbf{ 0.00}  
&         1 &   0.05
&         3 &   0.4  
\\ 
 6 & 3 &      540 &  0.16  
&       118 &  \textbf{ 0.03}  
&       2 &  0.16  
&       183 &  9.6  
\\ 
 7 & 4&   195,212 & 84.50  
&     83,731 &  \textbf{15.49}  
&    130,010 & 1802.49  
&    63 & 15.8  
\\ 
 8 & 5 &   390,717 & 255.64  
&    256,582 &  58.42  
&   24,588 & 585.07
&   30 & \textbf{41.28} 
\\ 
\hline 

\end{tabular}}
\end{center}
\end{table}

Results of our experiments are presented in Table~\ref{t:t1}. 
Our BC decomposition performs better than the  $Occs$ decomposition,
both in runtime and in  number of backtracks needed by Ilog Solver or CPLEX. 
CPLEX is slower per node than Ilog Solver. However,
CPLEX usually requires fewer backtracks compared to ILOG Solver.
Interestingly CPLEX performs well with the BC decomposition.
The time to explore each node is large, reflecting 
the size of decomposition, but the number of search nodes
explored is small. We conjecture that integer linear
programming methods like CPLEX will perform in a similar way with
other decompositions of global constraints which 
do not hinder propagation (e.g. the decompositions we have
proposed for \alldiff and \gcc).
Finally, the best results here
are comparable with 
those for the $\AtMostNValue$ bounds consistency
propagator in~\cite{bhhkwconstraint2006}.

\myOmit{
\subsection {Random binary CSP problems}
We also reproduced the set of experiments on random binary CSP problems from ~\cite{bhhkwconstraint2006}.
These problems can be described by four parameters. The number of variables $n$, the domain size $d$,
the number of binary constraints $m$ and the number of forbidden tuples in each binary constraint.  
The first three classes are hard problems at the 
phase transition in satisfiability. 
The last two classes are under-constrained problems.
We add a single $\AtMostNValue$ constraint over all variables to bound the number of values $N$ that can be used in a solution.

As in \cite{bhhkwconstraint2006}, 
we generated 500 instances for each of the following 5 classes:
\begin{itemize}
	\item class A : $n = 100, d = 10, m = 250, t = 52, N = 8$
	\item  class B : $n = 50, d = 15, m = 120, t = 116, N = 6$
	\item  class C : $n = 40, d = 20, m = 80, t = 240, N = 6$
	\item class D : $n = 200, d = 15, m = 600, t = 85, N = 8$
	\item class E : $n = 60, d = 30, m = 150, t = 350, N = 6$
\end{itemize}

All instances are solved using the minimum domain variable ordering heuristic, a lexicographical value 
ordering and a timeout of $600$ seconds.  We use the same decompositions of
the \AtMostNValue constraint as in the experiments with the dominating 
set of the Queen's graph. Results are given in Table~\ref{t:t2}.  
On classes $A,B,C$ (hard problems), our decomposition  is faster than
the other two decompositions and solves more instances whatever we use
BC or RC
. 
On classes $D,E$ (under-constrained problems),  enforcing BC 
on our decomposition
does not prune 
the search space enough. This leads to a high number of backtracks and a
significant slow down. Our decomposition with RC 
is again better than the other decompositions.

\begin{table}[tb]
\begin{center}
{\small
\caption{\label{t:t2} Randomly generated binary CSPs with an $\AtMostNValue$ constraint. 
For each class we give two lines of results. Line 1:
number of instances solved in 600 sec (\#solved),  average backtracks on solved
instances (\#bt),  average time on solved instances (time). Line 2: 
number of instances solved by all methods,  
average backtracks and time on these instances. 
}
\begin{tabular}{|  cr |rrr|rrr|rrr|rrr|}
\hline
$$ 
&&\multicolumn {3}{|c|}{$Occs$}
&\multicolumn {3}{|c|}{$Occs_{gcc}$}
&\multicolumn {3}{|c|}{$Pyramid_{BC}$}
&\multicolumn {3}{|c|}{$Pyramid_{RC}$}
 \\
\hline 
\hline
$Class$ 
&&\multicolumn {3}{|c|}{\#solved ~~ \#bt ~~ time}
&\multicolumn {3}{|c|}{\#solved ~~ \#bt ~~ time}
&\multicolumn {3}{|c|}{\#solved ~~ \#bt ~~ time}
&\multicolumn {3}{|c|}{\#solved ~~ \#bt ~~  time}\\
\hline 
 A  & total solved & 453 & 139,120 &  111.2 & 79 &   8,960 &  302.8 & \textbf{463}&  \textbf{ 168,929} & \textbf{  101.8} & {462}&  { 148,673} & {  105.7} \\ 
&  solved by all & 79 & 8,960 & 7.1 & 79 & 8,960 &  302.8& 79 &   \textbf{   9,104} & \textbf{    5.7}&  79 &  {   8,739} & {    6.3} \\ 
\hline 
B  & total solved & 473 & 228,757 & 113.5 & 125 &  37,377 &  292.9 & \textbf{492}&  \textbf{ 224,862} & \textbf{ 89.0} & {491}& { 235,715} & {   94.9} \\ 
& solved by all & 125 & 7,377 & 17.6 & 125 & 37,377 &  292.9 &125 & \textbf{  32,810} & \textbf{   10.9} &125 & {  32,110} & {   12.2} \\ 
\hline 
 C  & total solved & 479 & 233,341 & 110.3 & 156 &  37,242 & 290.3 & \textbf{492}&  \textbf{ 234,915} & \textbf{   79.5} & {490}& { 224,802} & {   84.2} \\ 
& solved by all & 156 & 37242 &  16.4 & 156 &   37,242 & 290.3 &  156 & \textbf{  32,184} & \textbf{   9.7} &  156 & {  31,715} & {   11.1} \\ 
\hline 
\hline 
 D & total solved & 482 &   8,306 & 6.0 & 456 &    207 &   14.9 & {416}&  {  168,021} & {    24.2} & \textbf{489}&  \textbf{  13,776} & \textbf{    9.0} \\ 
 & solved by all & 391 & \textbf{    195} & \textbf{    0.2} & 391 & 195 & 13.1 &  391 & 145,534 & 14.9 & 391 & 690 & 0.4 \\ 
\hline 
 E & total solved & \textbf{500}&    331 & 0.3 & \textbf{500}&   331 & 5.1 & \textbf{500}&  {    4,252} & {    0.4} & \textbf{500}&  \textbf{    174} & \textbf{    0.1} \\ 
 & solved by all  & 500 & 331 & 0.3 & 500 & 331 & 5.1 & 500 & {    4,252} & {    0.4} & 500 & \textbf{    174} & \textbf{    0.1} \\
\hline 
 \multicolumn {2}{|c|} { TOTALS }& & & & & & & & & &&&\\ 
 \multicolumn {2}{|r|} {Total solved/tried}& \multicolumn {3}{|c|} { 2,387/2,500}& \multicolumn {3}{|c|} { 1,316/2,500}& \multicolumn {3}{|c|} {{2,363}  /2,500}& \multicolumn {3}{|c|} {\textbf{2,432}/2,500}\\ 
 \multicolumn {2}{|r|} {Avg time for solved}& \multicolumn {3}{|c|}{ 67.0} & \multicolumn {3}{|c|}{ 87.5}& \multicolumn {3}{|c|} {59.364} & \multicolumn {3}{|c|}{\textbf{ 58.0}} \\ 
 \multicolumn {2}{|r|} {Avg bts for solved}& \multicolumn {3}{|c|}{ 120,303} & \multicolumn {3}{|c|}{   8,700} & \multicolumn {3}{|c|} {163,473} &\multicolumn {3}{|c|}{\textbf{ 123,931}} \\ 
\hline 
\end{tabular}}
\end{center}
\end{table}

These experiments demonstrate that 
this new decomposition is not completely 
infeasible. Of course, if the toolkit contains a specialized 
BC propagator for the \nvalue constraint, we will probably do best
to use this. However, when the toolkit lacks such
a propagator (as is often the case), it may be reasonable
to try out our decomposition. 

 }
\section{Other related work}

Bessiere {\it et al.}
consider a number of different methods to
compute a lower bound on the number of 
values used by a set of variables
\cite{bhhkwconstraint2006}. 
One method is based on a simple linear relaxation
of the minimum hitting set problem. 
This gives a propagation algorithm that
achieves a level of consistency strictly
stronger than bound consistency on the
\nvalue constraint. 
Cheaper approximations
are also proposed based on greedy heuristics
and an approximation for the independence 
number of the interval graph due to Tur\'{a}n. 
Decompositions have been given for
a number of other global constraints. 
For example, Beldiceanu {\it et al.} identify conditions
under which global constraints specified
as automata can be decomposed into signature
and transition constraints without hindering
propagation \cite{bcdpconstraints05}. 
As a second example, many global constraints
can be decomposed using \roots and \range
which can themselves be 
propagated effectively 
using simple decompositions 
\cite{bhhkwijcai2005}. 
As a third example, 
the \regular and \grammar constraints can be decomposed
without hindering propagation
\cite{qwcp06,qwcp07}. 
As a fourth example,
decompositions of the \sequence constraint 
have been shown to be effective
\cite{bnqswcp07}. 
Most recently, we demonstrated that
the \alldiff and \gcc constraint
can be decomposed into simple primitive constraints
without hindering bound consistency propagation 
\cite{bknqwijcai09}. These decompositions
also introduced variables to count variables
using values in an interval. For example, the
decomposition of \alldiff ensures that
no interval has more variables taking 
values in the interval than the number of values in
the interval. 
Using a circuit complexity
lower bound, we also proved that
there is no polynomial sized SAT decomposition
of the \alldiff constraint (and therefore 
of its generalizations like \nvalue) on 
which unit propagation achieves
domain consistency \cite{bknwijcai09}.
Our use of ``pyramid'' variables is similar to the use of the
``partial sums'' variables in the encoding of the \sequence constraint
in~\cite{bnqswcp07}. This is related to the cumulative sums
computed in~\cite{vhprscp06}.

\section{Conclusions}

We have studied a number of decompositions
of the \nvalue constraint. 
We have shown that a simple decomposition
can simulate the bound consistency
propagator for \nvalue \cite{bcp01}
with comparable time
complexity but with a much greater
space complexity. This supports the conclusion
that the benefit of a global
propagator may often not be in 
saving time but in saving space. 
Our other theoretical contribution
is to show the first range consistency algorithm
for \nvalue, that runs in $O(nd^3)$ time and $O(nd^2)$ space.
These results are largely
interesting from a theoretical perspective. 
They help us understand the globality of 
global constraints. They highlight that
saving space may be one of the important
advantages provided by propagators for global
constraints. 
We have seen that the space complexity of decompositions
of many propagators equals the 
worst case time complexity (e.g. for the \alldiff, \gcc, \among, \lex, 
\regular, \grammar and \sequence constraints).
For global constraints like
\regular , the space complexity of the
decompositions does not appear to be that
problematic. 
However, for global constraints like \nvalue ,
the space complexity of the decompositions is onerous.
This space complexity
seems hard to avoid. For example, consider
encodings into satisfiability and unit propagation
as our inference method. As unit propagation is
linear in time in the size of the encoding, 
it is somewhat inevitable that the size of any encoding is 
the same as the worst-case time complexity of any
propagator that is being simulated.
One other benefit of these decompositions is that
they help us
explore the interface between constraint
and integer linear programming. For example, 
we saw that an integer programming solver 
performed relatively well with these decompositions. 

\textbf{Acknowledgements.}
{NICTA is funded by 
the Department of Broadband, 
Communications and the Digital Economy, and the 
ARC. 
Christian Bessiere is supported by ANR project  ANR-06-BLAN-0383-02,
and George Katsirelos by ANR UNLOC project: ANR 08-BLAN-0289-01.}
We thank Lanbo Zheng for experimental help. 

\bibliographystyle{splncs}



\end{document}